# Socially Assistive Robot in Sexual Health: Group and Individual Student-Robot Interaction Activities Promoting Disclosure, Learning and Positive Attitudes

Anna-Maria Velentza ˈ[0000-0002-1251-571X]ˈ [a,b*], Efthymia Kefalouka[a]

Nikolaos Fachantidis ˈ[0000-0002-8838-8091]ˈ [a,b]

[a]*Department, University, City, Country;* [b]*Department, University, City, Country*

[a]*School of Educational & Social Policies, University of Macedonia, GR;* [b] *Laboratory of Informatics and Robotics Applications in Education and Society (LIRES), University of Macedonia, GR*

*Corresponding author. E-mail(s): annamarakiv@gmail.com;
Contributing authors: ite21021@uom.edu.gr; nfachantidis@uom.edu.gr

# Socially Assistive Robot in Sexual Health: Group and Individual Student-Robot Interaction Activities Promoting Disclosure, Learning and Positive Attitudes.

**Abstract**

Comprehensive sex education (SE) in schools plays a vital role in establishing a profound link to and actively advocating for sexual health, aiming at educating children about sexual health, ethics, and behaviour. The implementation of SE in elementary schools can significantly transform students' attitudes and comprehension of sexual knowledge. However, teaching SE has been challenging at times due to students' beliefs, attitudes, and occasional shyness or emotional reservations. Socially assistive robots (SARs) sometimes are perceived as more trustworthy than humans, based on research showing that they are not anticipated as judgmental. Inspired by those evidences, this study aims to assess the success of a SAR as a facilitator for SE lessons for elementary school students. We conducted two experiments, a) a group activity in the school classroom where the Nao robot gave a SE lecture, and we evaluated how much information the students acquired from the lecture, and b) an individual activity where the students interacted 1:1 with the robot, and we evaluated their attitudes towards the subject of SE, and if they felt comfortable to ask SE related questions to the robot. Data based on given pre- and post-questionnaires and video annotations demonstrated that the SAR statistically significantly improved students' attitudes towards SE. Moreover, they addressed to the robot increased number of questions regarding SE and body parts in comparison with SE lesson taught by their teacher. The study also highlights the SAR characteristics that make them efficient to support SE, such as their embodiment and non-judgmental behavior. This study is unique in its focus on emphasizing the SAR's potential to support SE for elementary school students in a real class environment.

## 1. Introduction

**S**exual health is intrinsically connected to and actively promoted through comprehensive sex education in schools[1], [2]. Sex education (SE) is a comprehensive approach that covers a wide range of topics related to human sexuality, including anatomy, reproduction, emotional relationships, and contraception. Its aim is to educate children about sexual health, ethics, and behaviour, while also raising awareness about social issues such as gender discrimination. [3]. The implementation of SE in elementary schools brings about a significant transformation in the way students perceive and comprehend sexual knowledge and attitudes [4].

There are many important factors to effectively implement SE programs in ages 9-12. Some of them cannot be manipulated (i.e., kids' personal characteristics, family, and school values), while others focus on how to support and guide teachers[5]. Teachers' personal knowledge about the students can prove beneficial for SE[6], but on the other hand, some students seem to feel more comfortable discussing SE matters with external educators [7].

Aguilar Alonso et al [5] address important factors that promote SE programs. Students 9-12 should be involved in the interaction [8],[9],[6], while the program should be adapted for acceptability and appropriateness based on students' needs [6], [10]. Moreover, it is important for children to feel safe and secure[9], [6], [10], represented and included [9]. Fun-play, active, and embodied seem to promote students' learning regarding SE matters [11].

There is a rapid use of technology and intelligent solutions in education, including the use of SAR to support learning[12], [13], due to their ability to form social relationships[14], [15], [16] and be trusted by the kids[14], [17]. Although technological solutions such as tablets, digital media, the web, or social media have been deployed for SE[11], [18], to the best of our knowledge there is no extensive use of SAR for SE.

This paper aims to fill this gap in the literature, by designing and implementing SE lessons with the aid of a SAR in a real classroom environment. Despite getting informed by various sources about sex, there are still many challenges when teaching the subject[19], while young people still experience emotions of embarrassment and shame [19] and thus we need to explore new ways to make them feel comfortable when being informed about it. This study assesses the success of the SAR as SE facilitator in terms of how much the elementary school students who participated in the study remembered from the lesson, how enjoyable the lesson was, and by evaluating qualitatively and quantitative the questions they asked the robot when they interacted 1:1. The study took place in a Greek school. SE was mandatory included in the curriculum from kindergarten to high school in September 2021.

The most important findings of our study are that: a) the interaction with a SAR significantly improves the students' attitudes towards SE; b) the introduction of the use of SAR, significantly increases the number of questions students asks regarding SE and body parts. To the best of our knowledge, there has been no other research work so far covering the interaction of SAR with elementary school students in both group and individual activities, highlighting the SAR characteristics that can be used to support SE.

## 2. Related Work

The widespread adoption and popularity of technology have paved the way for digital media interventions in SE, offering a promising avenue for promoting sexual health knowledge. This is achievable through online platforms, such as eHealth, and mobile applications (i.e., mHealth), which grant individuals, particularly young people, the benefits of privacy and anonymity. Blended learning programs emerge as a favorable approach, effectively amalgamating the strengths of in-person and digital interventions, as noted by Lameiras, et al,2021 [18]. The use of digital interventions in school settings, both within and beyond the classroom, presents intriguing possibilities due to their heightened flexibility in addressing diverse learning needs and surpassing traditional face-to-face methods. These digital interventions also excel in their capacity for customization, interactivity, and the provision of a secure, regulated, and familiar environment for disseminating sexual health knowledge and skills [20]. The escalating utilization of the internet and digital media for SE is driven by the unique advantages they offer, including portability, anonymity, informality, and personalized responses, distinguishing them from conventional sources of sex education[21].

Prior to utilizing a SAR coordinated activity for aiding teachers in SE, we conducted thorough bibliographic research on the robots' characteristics that can improve this interaction and also to answer, 'Why socially assistive robots?'. Our attention was focused on SAR's appearance (i.e., perceived gender), human perceptions and attitudes after human-robot interaction, and prior relevant uses.

Trust is very important when it comes to sensitive matters such as SE. According to previous studies[14] children are more inclined to form attachments with robots compared to adults. Their research demonstrated that over 50% of the children were willing to converse with the robot, and nearly 50% of them even felt comfortable sharing their secrets with it[14]. Based on another research, kindergarten kids are equally likely to trust both an adult and a social robot when it comes to sharing their secrets [22]. SARs' physical embodiment plays a significant role in fostering a trusting relationship between humans and robots [15], [16], contributing also to the development of comfort [23]. Moreover, humans can grow an attachment to a SAR even from a short, but meaningful interaction[24]. In another study, students were asked if they prefer to be taught by a human or a robot teacher and those who argued in favor of the robot highlighted that it wouldn't punish them or get angry, while at the same time, it would listen to them, and let them have their own way [17]. Moreover, consumer studies showed that robots are perceived as non-judgmental, and customers feel less judged by a robot than by a human in situations that might be embarrassing, such as acquiring medication to treat a sexually transmitted disease or being confronted with their mistakes by a frontline employee[25]. Interacting with service robots can reduce anticipated embarrassment based on consumers' evaluations because robots are not able to make moral or social judgments[26].

Social studies on teacher-student gender, support the idea that we tend to prefer people who are similar to ourselves, whether it's in terms of gender or culture[27]. In the ABOT dataset, humanoid robots were mostly perceived as young adults with a gender that was mostly neutral or masculine[28]. Moreover, SARs have been proven beneficial in delivering knowledge to K-12 students [12], [13]. It is also worth mentioning that K-12 students were able to follow orders and successfully complete educational tasks after having 1:1 sessions with the NAO robot[29].

Regarding the use of SARs' as mental health facilitators, research findings indicate that robot interventions for children have the potential to improve their mental health outcomes, including alleviating distress and enhancing positive emotions[30]. However, there is a need to strengthen the quality of evidence to identify which specific types of robotic interventions would be most effective and feasible to be implemented in pediatric mental healthcare[30].

SARs' have also been deployed to provide SE. A study examined the effectiveness of a psychological intervention program using a smart robot to promote good sexual care among primary school children. The experimental group received the program, while the control group did not. The program had a positive impact on the children's sexual care and increased their knowledge about appropriate touch. However, further parameters need to be tested, since any intervention is considered better than none [31]. Moreover, it has been proposed the participation of sex robots in child sexual abuse treatment, with experts, physicians, and therapists providing heterogeneous feedback on the idea[32].

## 3. Present study

Given the research results described in the last section it is clear that SARs have some core characteristics, i.e., ability to be trusted by young kids[22], grow attachment[24], and perceived as non-judgmental[17], [25], [26] that can be useful to assist teachers deliver a sex education course. Moreover, they can effectively impart students' knowledge when they deliver a course[12], [13], [29]. Moreover, a primary facilitation of SARs delivering a sex education program gave very promising results[31]. Moreover, we believe that the robot's perceived young age and neutral gender [28], will also enhance the students-robot interaction during the SE[27].

It is important to enhance students' gained knowledge, but most importantly, sex education is about building trust, respect, and accept and understand their body, and safe practices. Thus, we designed a lecture in close collaboration with the class teacher as an introduction to get to know their bodies and we utilized both a group and an individual activity to also test how peer pressure affect them to express their feelings and ask more questions [33]. We evaluated students' attitudes and gained knowledge based on pre and post-tests, video annotations, the number of questions they asked regarding SE during the activities and their responses to short interview questions after the activity.

### 3.1 Hypothesis

H1: We expect that students will increase their knowledge acquisition after having a lesson with the robot-tutor[11], [12]. Previous studies have shown that SARs can successfully support health education[34]. If this hypothesis will be proven right, we expect higher scores in the post-test knowledge quiz in comparison with the pre-test.

H2: Elementary school students seem to feel comfortable interacting with SAR [35] and they improve their attitudes towards the use of robots in education and also their attitudes towards the taught subject after having lectures with them[36]. Most of the studies focus on stem topics, and thus we are interested to investigate if the interaction with a SAR can improve their attitudes towards SE. We expect that students will have higher scores in the attitudes post-test questionnaires.

H3: We expect students to address more questions to the robot during the 1:1 interaction activity, in comparison with their teacher. Students seem to feel more comfortable discussing SE matters with external educators [7], the robot will not judge them[17],[25],[26], and also during the 1:1 activity, they will not feel the peer pressure that they will be judged by their classmates[33].

## 4. Experimental studies

We conducted two experiments taking place in the same day, following the sequence depicted in Fig.1 that will be explained thoroughly on the next chapters.

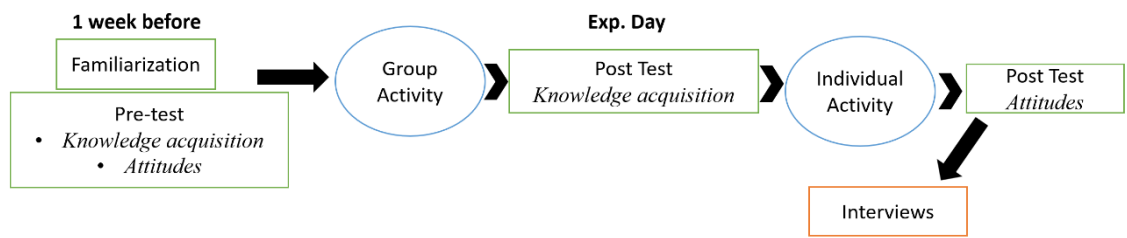

**Fig. 1** Step by step study design for both Experiment 1 & 2

## 4.1 Experiment I: Sex education with robot-tutor (group activity)

In Experiment I, the robot-tutor taught students about SE in front of the whole class, mimicking the standard teaching activity, performed by human-tutors. The classroom teacher had performed lecture regarding sexual health two months before and she had mentioned that none of the students expressed the interest to ask anything sex health related during the class, neither when they privately asked them if they had any questions. The purpose of Experiment I is to set the bar for a group class activity with the robot-tutor to stress the students' attitudes and gained knowledge after being taught by the robot.

### *4.1.1 Participants*

The participants were 11 students, 3 boys and 8 girls, 9 years old, studying on the 3rd grade of a rural elementary School in Imathia, Greece, in the school year 2021-2022. They all had normal or corrected to normal vision and hearing and they were native speakers of the used language. The Ethics Committee of the University of Macedonia approved and supervised the experimental procedure, and both parents and students where thoroughly informed about the goals of the study and signed the consent forms.

### *4.1.2 Design*

The experimental design was within group. Extensive bibliographic research indicates that half of SE programs involved focus on understanding body-parts, their functions and the human reproduction[5]. The lecture taught by the robot was designed in collaboration with the classrooms' teacher, one independent teacher and one psychologist. The robot presented 10 body-parts to the students, including 4 external body-parts (eyes, nose, mouth, nose), 4 internal (heart, brain, digestive system, lungs), and in two pairs, the male and female reproductive system, explaining their function, based on their age i.e., 'Lungs are two large sacs located inside our chest. The left is a little smaller than the right because it makes room for our heart. With the lungs we inhale and exhale air' or 'A girl has these genitals. They are divided into internal and external, depending on their location inside or outside the body. The internal body-parts are the vagina, ovaries and fallopian tubes. The outer ones are called the vulva'. We draw, printed and laminated representations of the forementioned body-parts, to help students create mental representations[37], emphasizing in having a cartoonish appearance[38] as shown in Fig.2. For the same purpose and to avoid gender biases, we used a luminated 2D representation of the Nao robot, shown in Fig.3.

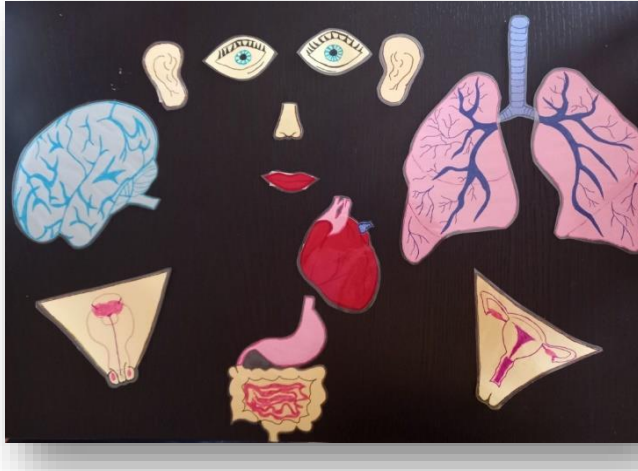

**Fig. 2** Laminated pictures of human body-parts.

The Nao robot was programmed using the Choreographer software platform, whereby automation was achieved through the integration of body part recognition and tactile sensor interaction. Specifically, the programming involved the identification of body parts and the utilization of tactile sensors located on the robot's head and hand to facilitate automated actions and responses. Moreover, we followed the recommendations of [39] to improve the students' attention span without distracting them. The voice speed and depth were finalized after a pilot study with same-age students, listening to short sentences with a variety of speeds and depths. Moreover, there were made specific language adaptations for all the words to sound clear and with the correct Greek pronunciation.

The robot was also programmed to perform a dance that was used as ice-breaker, synchronized with the music of the kids' Greek song 'head, shoulders, knees and feet', by pointing to the corresponding body-part along with the music. The description for each body-part had been saved as a separate behaviour and the robot's body language had been set to the natural mode based on the content of the lecture, as proposed by[40]. The intro, outro, feedback, and small talk phrases had also been pre-programmed. The volume had also been tested in classroom environment.

To evaluate students' *knowledge acquisition*, in close collaboration with the class teacher, we designed a learning questionnaire containing 13 questions based on the lecture's content which was given to the students before and after the lecture. There were 9 multiple choice questions with 4 equally possible answers and only one correct (i.e., 'the lung's basic function is to help us' 1. eat, 2. breath, 3. think, 4. sleep), 3 true/false (i.e., 'the heart is a very powerful muscle that beats even when we sleep' True/False) and one open question.

The whole procedure was filmed, and the video annotations served as additional methodological tools for data collection for reliable conclusions, combining qualitative and quantitative data sources[41].

### *4.1.3 Procedure*

One week before the experiment, the students had been informed by both their teacher and the second author regarding the activity. They showed them videos and photos of the robot, serving also as a familiarization activity as proposed by [42], and at the same time to avoid expectation biases from the robot's behavior [40]. Moreover, they filled in and submitted the pre-questionnaire, as shown in Fig.1.

The experiment took place at 10.00am on a weekday, during the teaching hours and lasted for 22min. The students entered their classroom, and the robot was already standing in the middle of the teacher's desk to be visible to everyone as shown in Fig.3. The class teacher was standing at the back of the room, while the second author, mentioned as researcher-also a teacher-next to the desk. A puppet show set up which is normally used by students during free hours was standing open next to the desk, to hide the first author who was serving as The Wizard of Oz, and the equipment (laptop, extra cables, etc.). The desks were settled two to three meters from the desk to give the students space to dance during the ice-breaker activity.

For the welcome state, the researcher repeated the procedure and the reason why the robot was there together with some basic safety rules while asking the students to remain quiet. The robot thanked the experimenter, introduced itself, and started the ice-breaker activity, encouraging the students to dance to the lyrics of the song by following its lead. The class teacher was frequently using the same ice-breaker activity, and thus the kids were familiar with the song and the choreography. Subsequently, the robot made an introduction regarding SE and body-parts. One by one, the students approached the desk, picked up one of the laminated body-part pictures of their choice, and show it to the robot. The robot was giving relevant information and then the students moved to the robot laminated figure to pin the body-part in the right place, as shown in Fig3.

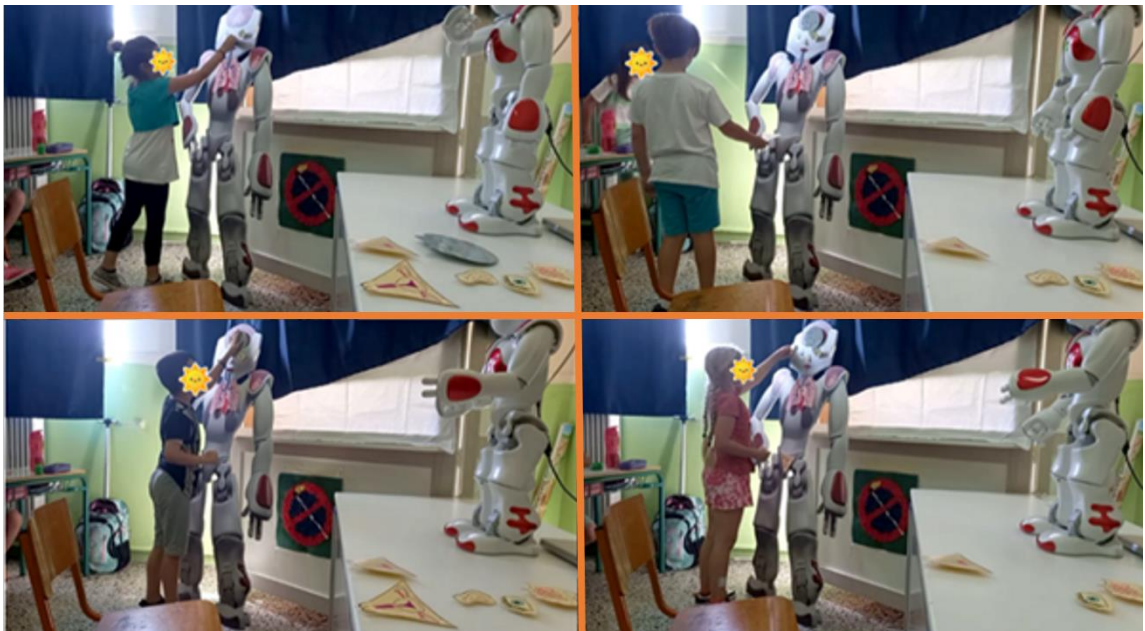

**Fig. 3** Group Activity. Students individually pin their chosen body-part on the laminated figure of the robot in front of their classmates.

Each body-part was represented by one picture, except for the eyes and ears that were two (Fig.2). At the end of the activity, the robot thanked the students for their participation and the students followed their teacher to the pc room to fill in the knowledge acquisition questionnaire(post-test), as shown in Fig.4 which also thoroughly depicts the procedure.

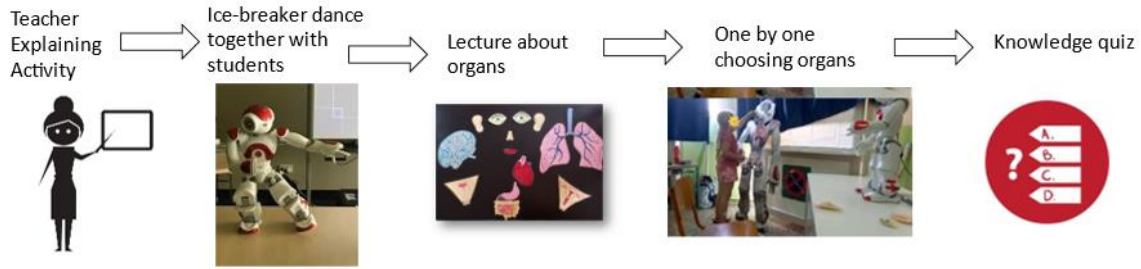

**Fig. 4** Group Activity Experimental Procedure.

*4.1.4 Data Analysis*

The maximum achieved score in the learning quiz was 17. We computed the mean and standard deviation for each student within the categories of pre-test and post-test. After using the Kolmogorov-Smirnov test, we analyzed the pre-and post-test results through a t-test.

*4.1.5 Results*

The students improved their knowledge based on the lesson between the pre-test (*MV*=12.09, *SD*=2.81) and the post-test (*MV*=13.27, *SD*=2.49), but not statistically significant, as indicated by the t-test at t(10)=1.633,p=.134,d=2.401. It is worth mentioning that 9 out of 11 students improved their scores on the test, while 2 students performed poorly in the post-test in comparison with the pre-test. The score of one student (S8) out of the 2 who performed poorly in the post-test decreased by 3 points, and thus we assume that the student did not pay attention while submitting the post-test. After excluding that one student from the analysis, the students' scores in the pre-test were *MV*=11.6, *SD*=2.41, and in the post-test were *MV*=13.2, *SD*=2.61.

The students preferred to pick up external and internal body-parts and only the last student picked up reproductive parts because there were no other parts left on the table, as shown in Table 4. It is worth mentioning that S10 felt awkward by this choice, lowered his head, and completed the procedure faster than his classmates, as shown by the camera footage.

*4.1.6 Discussion*

Most of the students improved their learning scores, however, the difference between the pre and the post-test is not significant (H1). Evidences from college students show that during the first lesson with a robot tutor, students did not significantly improve their learning scores due to the surprise effect -paying more attention to the robot itself and not to the storytelling-, however, they increased their long-term learning and motivation[43]. To eliminate the effect we used the video familiarization phase as proposed by [42], yet, we still consider the existence of surprise effect since it was the first time the students interacted with an actual robot.

*4.2 Experiment II: Robot- Students individual activity*

Based on the research results described in the introduction section, it is crucial for kids to feel comfortable, included, and safe. The existence of a SAR can strongly affect students' attitudes towards sex education due to its social characteristics, but this was not tested in an actual classroom environment. It is worth mentioning that during the SE with their teacher or the external educator, students did not ask any questions, even when they were encouraged privately to do so. Experiment II took place right after Experiment I.

### 4.2.1 Participants

The participants were the same students with Experiment I.

### 4.2.2 Design

The students interacted with the robot via The Wizard of Oz approach. The first author who is also a psychologist served as the wizard, due to the sensitivity of the information and the topic. There have been pre-coded responses and behaviours, while when there were new things that the wizard had to write, the robot asked the students to be patient with it, as it will reply immediately. To avoid being noticed by clicking on the keyboard, the wizard used a soft keyboard extension.

To evaluate the students' attitudes towards SE, the use of robots in general, and the use of robots in SE, we asked them to submit one pre and one post-questionnaire with 26 Likert scale questions. The first part had 9 questions regarding their attitudes towards SE and was based on [44] questionnaire, (i.e.,'I'm not interested to learn about SE','I feel comfortable discussing with my friends about sex education'). The second part contained 7 statements describing what did they expect to feel if they have a lecture with a robot-tutor regarding SE (pre-test)/ how they felt during the lecture with the robot (post-test), i.e., curiosity, interest, the need to pay more attention, the urge to make fun of the course. The third part was based on [45], with 4 questions regarding their attitudes towards robots in general (i.e., 'Robots are dangerous', 'Robots are likable')[45]. The fourth part had 3 questions, evaluating if the students think that with the use of a social robot, the course will become more interesting, easily understandable, pleasant[36], and lastly, we incorporated 3 questions asking the students to evaluate if the robot would/ did a) help them concentrate better, b) enhance them to ask more questions, and c) in case they feel uncomfortable will make them feel more comfortable.

### 4.2.3 Procedure

A chair was placed between the robot that stayed in the same position at the desk, but this time facing the chair and the picture of the robot, as shown in Fig.4. The wizard was seated hidden behind the doll's house. The students entered the room one by one and were instructed to sit on the chair and follow the robot's instructions. The teacher was calling the students by their names after they entered the class, so the wizard listened to the name and make the robot address them with it when greeting them. The students were informed that they were able to stop and exit the room at any time that they wanted and also that the teacher will wait for them outside the room.

In the first part, the robot greeted the students by their names and explained that they will do an activity together. Then, the robot instructed them to show it three body-parts of their choice -from the ones on the desk that they also used for the group activity- to give them more information about them and then to pin them on the robot's laminated picture. In the second part, the robot encouraged the students to ask whatever they want regarding the body-parts mentioning that 'now that your teacher is not here and is not able to listen to you, is anything that you would like to ask me?'. When the students did not have anything else to ask or when they asked the robot if they can leave, the robot said goodbye and thanked them for their participation. After they left the room, the experimenter asked them personally if they liked the activity and what do they think about the robot, without expressing any opinion. In the end, the students submitted the attitudes post-test.

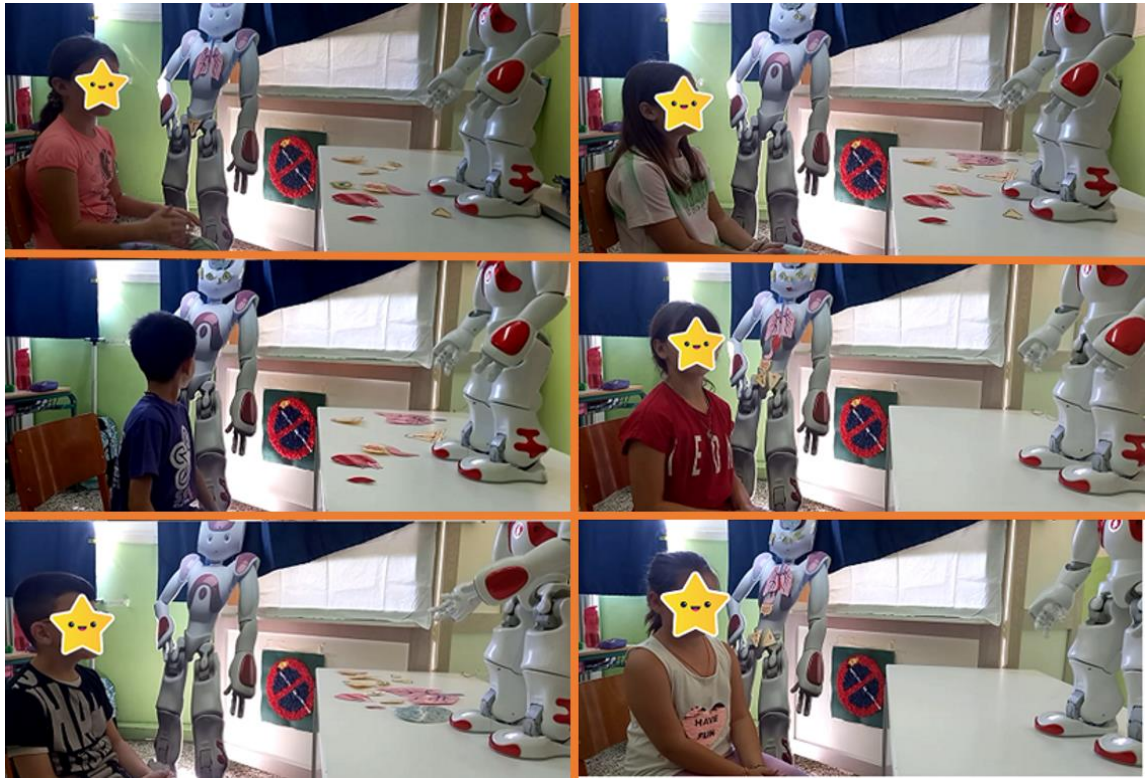

**Figure 4.** Student-robot individual activity

## *4.2.4 Data Analysis*

We first calculated the Sum, *MV*, and *SD*, per student for the attitudes' questions and then after applying the Kolmogorov-Smirnov test, we compared the pre-and post-test results with a one-tailed t-test. We wrote down and then calculated the percentage of students who asked about the body-parts, after separating the parts into three categories, 1) internal (brain, heart, lunges, etc.), 2) external (eyes, ears, legs), and 3) reproductive, either male or female. For the video analysis, we used the ELAN software[46], to evaluate the students' posture, and facial expressions during the interaction with the robot.

## *4.2.5 Results*

The students' attitudes toward both SE and the use of robots in education also improved after the lecture with the robot. Regarding the students' attitudes towards SE, the sum in the pre-test was 218 (*MV*=21.27, *SD*=3.26), and in the post-test 283 (*MV*=25.73, *SD*=4.05), which is statistically significant as indicated by *t*(10)=-2.84, *p*=.005,*d*=65.

The students' emotions (2$^{nd}$ part) remained relatively similar when they were taught about SE from the robot-tutor (Sum=182, *MV*=18.27, *SD*=4.24) in comparison with the human-tutor (Sum=177, *MV*=17.72, *SD*=4.82), but the difference is not statistically significant (t(10)=-0.28, p=.39,d=5).

Students remained similarly positive towards robots in general (3rd part) before (Sum=92, *MV*=8.36, *SD*=3.2) and after (Sum=118, *MV*=10.73, *SD*=2.26), indicated by t(10)=-1.72,p=.051,d=26.

When the students were asked to evaluate if they think that the presence of a robot can improve a SE lecture (4th and 5th part), they statistically significantly supported the idea after interacting with the robot (Sum=261,*MV*=23.73,*SD*=3.6) in comparison with before (Sum=191,*MV*=17.36,*SD*=3.29), as indicated by t(10)=-4.32, p=<.001,d=70.

The students interacted with the robot for an average of 6.47 sec and asked the robot about 73 body-parts. 45.21% (N=33) body-parts were internal, 39.73% (N=29) were external, and 15.1% (N=11) were reproductive body-parts. During the first phase, students picked up and asked for more information about 14 internal (41.17%), 17 external (50%), and 3 (8.82%) reproductive ones. During the second phase, after the robot encouraged the students to ask more questions if they wanted to, 3 students did not want to ask more, while the rest of the students' questions were about 19 internal (48.72%), 12 external (30.77%), and 8 reproductive ones (20.51%), as shown in Table 1.

**Table 1** Students' chosen body parts through all activities. G reprod stand for girls' reproductive parts and B. reprod correspondingly for boys'. Last column: interaction duration in minutes for individual activity.

| St. | Gender | Group Activity | 1st phase 1:1 | 2nd phase 1:1 | Duration (min) |
|---|---|---|---|---|---|
| S1 | G | Heart | Lungs/heart/eyes | Heart/lungs | 7.30 |
| S2 | G | Nose | Brain/G reprod./lungs | G reprod. ×3 | 8.23 |
| SE | G | Ear | Eye/ear/brain | Heart/ stomach/mouth/nose/ear | 5.86 |
| S4 | G | Eye | Eye/ear/nose | No | 3.86 |
| S5 | B | Eye | G reprod/ mouth/ heart | Heart/lungs/brain | 6.69 |
| S6 | G | Ear | Mouth/heart/ G.reprod. | brain/eye/lungs/heart/mouth/ear/ digestive/nose/B.reprod | 9.04 |
| S7 | G | Mouth | Ear/eye/nose | Brain/lungs/heart×2/ear/digestive/ G.reprod./B.repro. | 5.7 |
| S8 | G | Digestive | Heart/eye/brain/ear | No | 3.48 |
| S9 | B | Brain | Heart/lungs/digestive | Eye/nose/G.reprod./ear/brain/mouth/ B.reprod. | 10.8 |
| S10 | G | Lungs | Heart/ear/eye | Legs/elbow | 5.92 |
| S11 | B | G.reprod. | Brain/eye/ear | No | 4.32 |

Students who avoided choosing a reproductive body-part in the group activity chose it for the 1:1 interaction with the robot (S2,S5,S6)- and also students who asked about reproductive body-parts after the robot encouraged them to do so by telling them that their teacher cannot listen to what they are asking (S7,S9). S6 asked about the opposite gender's reproductive body-parts after the encouragement, while S2, asked three different questions regarding the reproductive body-parts of her gender.

Based on the camera annotations, the students smoothly interacted with the robot concentrate, and paid attention to the robot's instructions. Most of the time they were making eye contact with the robot, and they shifted their gaze to the body-parts or to the laminated robot figure at approximately 30% of their interaction time.

S5 and S7 were the ones who enjoyed the activity the most, spending up to 50% of their interaction time smiling at the robot, and they looked excited.

S6 and S8 at the end of the activity asked the robot personal questions (i.e., what is your name). S6 was an interesting case since at the beginning she refused to participate, informing the teacher that she is afraid of the robot. The teacher encouraged her by telling her that she can give it a try and that she will be outside the door, and S6 agreed. At the beginning of the interaction, she moved the chair further away from the robot, after 2.5 sec of interaction she moved the chair closer, and she ended up having the second-highest interaction time during the activity.

Excluding the time that the students were moving to pin the chosen body-parts on the robot, S3,4,8,10,11 spend the whole interaction being seated on the chair, S5 moved the chair to be seated closer to the robot, similarly with S2 and S7 who after the first part of the interaction stood up closer to the robot, touching the desk.

S1 was the only one who was standing behind the chair during the interaction. After the interaction, she confirmed that there were times during the interaction when she felt fear, without being able to explain why, even though she finished the activity and actively asked questions about more body-parts.

Moreover, when they were asked after the activity how they felt interacting with the robot, almost all replied positively mentioning that they felt 'good', 'interesting', 'ok', 'wow' 'I liked it', 'Can we please do it again?' and described the robot as 'good', 'nice', 'cool', 'fine'. Besides, they wanted to explain to the teacher how they interacted with the robot, and they got excited by saying 'then it replied', 'It was looking at me'. 5 students also asked if they are going to see the robot again, while 6 of them had questions on how the robot was operating.

Finally, it is worth mentioning that all students followed the robot's instructions, and no one tried to explore the room or revealed the wizard.

### *4.2.6 Discussion*

The outcome of this analysis is that students felt more comfortable asking the robot more questions than their teacher, and they increased the number of questions regarding reproductive body-parts when they were encouraged by the robot to ask more questions. However, some of the questions seem to be formed as a test of the robot's abilities and/or intelligence. 2 students (S6,8) wanted to display all the body-parts.

Although students had positive attitudes towards the use of robots in general, even before the experiment (pre-test scores), they statistically significantly scored higher when they were evaluating the usefulness of a robot-tutor in SE, and they also improved their attitudes towards SE in general after interacting with the robot-tutor.

All students followed the robot's instructions and successfully completed the activity. 10/11 students enjoyed the interaction and positively evaluated the robot when they were asked by the experimenter, except for S1, who expressed fear. In elementary school, having a varied vocabulary to express different emotions becomes important in distinguishing individual levels of emotional comprehension[47], and thus students' individual characteristics and personality traits need further investigation.

### 5. General Discussion and Conclusions

Our study focused in evaluating the effectiveness of using a socially assistive robot (SAR) as a facilitator for sex education (SE) lessons in an elementary school setting. We were interested mainly in how the robot can improve the students' knowledge about body-parts, as well as in the attitudes of the students regarding the interaction with the robot and SE in general. Moreover, we were investing in testing the effectiveness of SAR, based on our bibliographic research pointing out their ability to make humans trust them[22], develop attachment [14], [15] collaborate with them[16], and transfer knowledge [12], [13]. Two experiments were conducted: one involving group activities with the Nao robot delivering SE content, and another with individual interactions between students and the SAR.

Following the evidence listed in related work, we have conducted two real classroom experiments with a robot-tutor; one group and one individual, 1:1 the robot with each student. We investigated the students' knowledge acquisition from the group teaching activity with pre-and post-tests. While most students improved their learning scores after interacting with a robot-tutor, the difference was not significant (H1).

Moreover, we evaluated the students' attitudes with the use of pre-and post-tests and with video annotations during the individual activity, where we also evaluated the number and the type of questions the students addressed to the robot. Students felt more comfortable asking the robot questions (H3), and their attitudes toward SE improved after interacting with the robot-tutor, indicating that the use of robots in SE may positively influence students' perceptions of the subject (H2). However, some questions seemed to be a test of the robot's abilities or intelligence. All students successfully completed the activity, and most enjoyed the interaction with the robot, except for one student who expressed fear. Students who initially avoided discussing reproductive body parts in group activities became more engaged and willing to ask questions when interacting individually with the robot. Encouragement from the robot seems to have played a significant role in this shift. Further investigation is needed to explore students' individual characteristics and personality traits.

Most students had positive attitudes towards the robot, describing it as "good," "nice," and "cool." They also expressed a desire to interact with the robot again and had questions about its operation. Additionally, most students enjoyed the interaction and expressed positive feelings. However, one student experienced fear during the interaction, highlighting the importance of understanding individual emotional responses to robots in educational settings.

Hence, overall, our results suggest that the utilization of the SAR to support SE is successful. Although students did not statistically significantly improve their learning outcome, in SE it is more important to feel safe, understand, and respect. Students improved their attitudes, respected the robot, and increased their curiosity by asking more questions in comparison with the activities with their teacher, and the external educator.

On the other hand, our results clearly demonstrate that the use of SAR can be beneficial to support SE, the Nao robot is an expensive tool, and it needs specialized knowledge to be programmed and handled. Thus, we suggest future research to experiment with various intelligent agents of low-cost robots/agents with social appearance. Finally, one limitation of the study is that we used a binary approach (girl/boy) which does not fully account for gender diversity.

In conclusion, the study demonstrated that a SAR can be an effective tool for facilitating SE lessons in elementary schools. It increased engagement and positively influenced students' attitudes towards SE. However, further research is needed to understand individual emotional responses and personality traits in response to robot interactions.

## References

[1] A. Bourke, D. Boduszek, C. Kelleher, O. McBride, and K. Morgan, 'Sex education, first sex and sexual health outcomes in adulthood: findings from a nationally representative sexual health survey', *Sex Educ.*, vol. 14, no. 3, pp. 299–309, May 2014, doi: 10.1080/14681811.2014.887008.

[2] D. A. Rowe, J. Sinclair, K. Hirano, and J. Barbour, 'Let's Talk About Sex … Education', *Am. J. Sex. Educ.*, vol. 13, no. 2, pp. 205–216, Apr. 2018, doi: 10.1080/15546128.2018.1457462.

[3] K. K. Toor, 'A Study of the Attitude of Teachers, Parents and Adolescents towards Sex Education', *MIER J. Educ. Stud. Trends Pract.*, pp. 177–189, Nov. 2012, doi: 10.52634/mier/2012/v2/i2/1568.

[4] S.-J. Kim, J.-E. Lee, S.-H. Kim, and K.-A. Kang, 'The Effect of Sexual Education on Sex Knowledge & Attitude in Elementary School Students', *J. Korean Public Health Nurs.*, vol. 26, no. 3, pp. 389–403, 2012, doi: 10.5932/JKPHN.2012.26.3.389.

[5] R. Aguilar Alonso, K. Walsh, L. van Leent, and C. Moran, 'School-based relationships and sexuality education programmes in primary schools: contexts, mechanisms and outcomes', *Sex Educ.*, vol. 0, no. 0, pp. 1–20, Feb. 2023, doi: 10.1080/14681811.2023.2167816.

[6] S. Mason, 'Braving it out! An illuminative evaluation of the provision of sex and relationship education in two primary schools in England', *Sex Educ.*, vol. 10, no. 2, pp. 157–169, May 2010, doi: 10.1080/14681811003666366.

[7] P. Pound *et al.*, 'What is best practice in sex and relationship education? A synthesis of evidence, including stakeholders' views', *BMJ Open*, vol. 7, no. 5, p. e014791, May 2017, doi: 10.1136/bmjopen-2016-014791.

[8] H. Cahill *et al.*, 'An Integrative Approach to Evaluating the Implementation of Social and Emotional Learning and Gender-Based Violence Prevention Education', *Int. J. Emot. Educ.*, vol. 11, no. 1, pp. 135–152, Apr. 2019.

[9] H. Kamara, 'Development of a Comprehensive Sexual Education Curriculum for a Private, Independent Day School for Students Pre-K Through 8th Grade - ProQuest', Yale University ProQuest Dissertations Publishing, 2020.

[10] K. V. Newby and S. Mathieu-Chartier, 'Spring fever: process evaluation of a sex and relationships education programme for primary school pupils', *Sex Educ.*, vol. 18, no. 1, pp. 90–106, Jan. 2018, doi: 10.1080/14681811.2017.1392297.

[11] S. K. W. Chu *et al.*, 'Promoting Sex Education Among Teenagers Through an Interactive Game: Reasons for Success and Implications', *Games Health J.*, vol. 4, no. 3, pp. 168–174, Jun. 2015, doi: 10.1089/g4h.2014.0059.

[12] T. Belpaeme, J. Kennedy, A. Ramachandran, B. Scassellati, and F. Tanaka, 'Social robots for education: A review', *Sci. Robot.*, vol. 3, no. 21, Aug. 2018, doi: 10.1126/scirobotics.aat5954.

[13] R. van den Berghe, J. Verhagen, O. Oudgenoeg-Paz, S. van der Ven, and P. Leseman, 'Social Robots for Language Learning: A Review', *Rev. Educ. Res.*, vol. 89, no. 2, pp. 259–295, Apr. 2019, doi: 10.3102/0034654318821286.

[14] W. Barendregt, A. Paiva, A. Kappas, and A. Vasalou, 'Child-Robot Interaction: Social Bonding, Learning and Ethics', in *Workshop proceedings of Interaction Design and Children Conference IDC14*, 2014.

[15] A. Tapus, C. Tapus, and M. Mataric, 'The role of physical embodiment of a therapist robot for individuals with cognitive impairments', in *RO-MAN 2009 - The 18th IEEE International Symposium on Robot and Human Interactive Communication*, Sep. 2009, pp. 103–107. doi: 10.1109/ROMAN.2009.5326211.

[16] E. Deng, B. Mutlu, and M. J. Mataric, 'Embodiment in Socially Interactive Robots', *Found. Trends® Robot.*, vol. 7, no. 4, pp. 251–356, Jan. 2019, doi: 10.1561/2300000056.

[17] V. H. Y. Kwok, 'Robot vs. Human Teacher: Instruction in the Digital Age for ESL Learners', *Engl. Lang. Teach.*, vol. 8, no. 7, pp. 157–163, 2015.

[18]  M. Lameiras-Fernández, R. Martínez-Román, M. V. Carrera-Fernández, and Y. Rodríguez-Castro, 'Sex Education in the Spotlight: What Is Working? Systematic Review', *Int. J. Environ. Res. Public. Health*, vol. 18, no. 5, Art. no. 5, Jan. 2021, doi: 10.3390/ijerph18052555.

[19]  A. MacKenzie, N. Hedge, and P. Enslin, 'Sex Education: Challenges and Choices', *Br. J. Educ. Stud.*, vol. 65, no. 1, pp. 27–44, Jan. 2017, doi: 10.1080/00071005.2016.1232363.

[20]  A. Waling, A. Farrugia, and S. Fraser, 'Embarrassment, Shame, and Reassurance: Emotion and Young People's Access to Online Sexual Health Information', *Sex. Res. Soc. Policy*, vol. 20, no. 1, pp. 45–57, Mar. 2023, doi: 10.1007/s13178-021-00668-6.

[21]  L. Widman and et al., 'Full article: Feasibility, Acceptability, and Preliminary Efficacy of a Brief Online Sexual Health Program for Adolescents', *Journal of Sex Research*, vol. 57, no. 2, 2020, doi: https://doi.org/10.1080/00224499.2019.1630800.

[22]  L. Waldman and I. Amazon-Brown, 'New Digital Ways of Delivering Sex Education: A Practice Perspective', Feb. 2017, doi: 10.19088/1968-2017.104.

[23]  C. L. Bethel, M. R. Stevenson, and B. Scassellati, 'Secret-sharing: Interactions between a child, robot, and adult', in *2011 IEEE International Conference on Systems, Man, and Cybernetics*, Oct. 2011, pp. 2489–2494. doi: 10.1109/ICSMC.2011.6084051.

[24]  S. Reig, J. Forlizzi, and A. Steinfeld, 'Leveraging Robot Embodiment to Facilitate Trust and Smoothness', in *2019 14th ACM/IEEE International Conference on Human-Robot Interaction (HRI)*, Mar. 2019, pp. 742–744. doi: 10.1109/HRI.2019.8673226.

[25]  L. Robert, 'Personality in the Human Robot Interaction Literature: A Review and Brief Critique', Social Science Research Network, Rochester, NY, SSRN Scholarly Paper ID 3308191, Dec. 2018. Accessed: Apr. 07, 2021. [Online]. Available: https://papers.ssrn.com/abstract=3308191

[26]  J. Holthöwer and J. van Doorn, 'Robots do not judge: service robots can alleviate embarrassment in service encounters', *J. Acad. Mark. Sci.*, Apr. 2022, doi: 10.1007/s11747-022-00862-x.

[27]  V. Pitardi, J. Wirtz, S. Paluch, and W. H. Kunz, 'Service robots, agency and embarrassing service encounters', *J. Serv. Manag.*, vol. 33, no. 2, pp. 389–414, Jan. 2021, doi: 10.1108/JOSM-12-2020-0435.

[28]  Y. Fan *et al.*, 'Gender and cultural bias in student evaluations: Why representation matters', *PLOS ONE*, vol. 14, no. 2, p. e0209749, Feb. 2019, doi: 10.1371/journal.pone.0209749.

[29]  G. Perugia, S. Guidi, M. Bicchi, and O. Parlangeli, 'The Shape of Our Bias: Perceived Age and Gender in the Humanoid Robots of the ABOT Database', in *Proceedings of the 2022 ACM/IEEE International Conference on Human-Robot Interaction*, in HRI '22. Sapporo, Hokkaido, Japan: IEEE Press, Nov. 2022, pp. 110–119.

[30]  A. Jones and G. Castellano, 'Adaptive Robotic Tutors that Support Self-Regulated Learning: A Longer-Term Investigation with Primary School Children', *Int. J. Soc. Robot.*, vol. 10, no. 3, pp. 357–370, Jun. 2018, doi: 10.1007/s12369-017-0458-z.

[31]  K. Kabacińska, T. J. Prescott, and J. M. Robillard, 'Socially Assistive Robots as Mental Health Interventions for Children: A Scoping Review', *Int. J. Soc. Robot.*, vol. 13, no. 5, pp. 919–935, Aug. 2021, doi: 10.1007/s12369-020-00679-0.

[32]   M. Tahan, G. Afrooz, and J. Bolhari, 'The effectiveness of smart robot psychological intervention program on good sexual care for elementary school children', *Shenakht J. Psychol. Psychiatry*, vol. 7, no. 6, pp. 53–65, Jan. 2021, doi: 10.52547/shenakht.7.6.53.

[33]   C. Eichenberg, M. Khamis, and L. Hübner, 'The Attitudes of Therapists and Physicians on the Use of Sex Robots in Sexual Therapy: Online Survey and Interview Study', *J. Med. Internet Res.*, vol. 21, no. 8, p. e13853, Aug. 2019, doi: 10.2196/13853.

[34]   L. Bursztyn and R. Jensen, 'How Does Peer Pressure Affect Educational Investments?', *Q. J. Econ.*, vol. 130, no. 3, pp. 1329–1367, Aug. 2015, doi: 10.1093/qje/qjv021.

[35]   O. A. Blanson Henkemans *et al.*, 'Using a robot to personalise health education for children with diabetes type 1: A pilot study', *Patient Educ. Couns.*, vol. 92, no. 2, pp. 174–181, Aug. 2013, doi: 10.1016/j.pec.2013.04.012.

[36]   M. Á. Conde, C. Fernández, F. J. Rodríguez-Lera, F. J. Rodríguez-Sedano, V. Matellán, and F. J. García-Peñalvo, 'Analysing the attitude of students towards robots when lectured on programming by robotic or human teachers', in *Proceedings of the Fourth International Conference on Technological Ecosystems for Enhancing Multiculturality*, in TEEM '16. Salamanca, Spain: Association for Computing Machinery, Nov. 2016, pp. 59–65. doi: 10.1145/3012430.3012497.

[37]   A.-M. Velentza, S. Ioannidis, and N. Fachantidis, 'Service robot teaching assistant in school class- room', in *IEEE/RSJ International Conference on Intelligent Robots and Systems (IROS)*, Las Vegas, NV, USA, 2020.

[38]   A. Eitel and K. Scheiter, 'Picture or Text First? Explaining Sequence Effects when Learning with Pictures and Text', *Educ. Psychol. Rev.*, vol. 27, no. 1, pp. 153–180, Mar. 2015, doi: 10.1007/s10648-014-9264-4.

[39]   C. Eker and O. Karadeniz, 'The Effects of Educational Practice with Cartoons on Learning Outcomes', *Int. J. Humanit. Soc. Sci.*, Dec. 2014.

[40]   E. Lopez-Caudana, P. Ponce, N. Mazon, and G. Baltazar, 'Improving the attention span of elementary school children for physical education through an NAO robotics platform in developed countries', *Int. J. Interact. Des. Manuf. IJIDeM*, vol. 16, no. 2, pp. 657–675, Jun. 2022, doi: 10.1007/s12008-022-00851-y.

[41]   A.-M. Velentza, N. Fachantidis, and I. Lefkos, 'Human-robot interaction methodology: Robot teaching activity', *MethodsX*, vol. 9, p. 101866, Jan. 2022, doi: 10.1016/j.mex.2022.101866.

[42]   U. Östlund, L. Kidd, Y. Wengström, and N. Rowa-Dewar, 'Combining qualitative and quantitative research within mixed method research designs: A methodological review', *Int. J. Nurs. Stud.*, vol. 48, no. 3, pp. 369–383, Mar. 2011, doi: 10.1016/j.ijnurstu.2010.10.005.

[43]   C. de Jong, R. Kühne, J. Peter, C. L. V. Straten, and A. Barco, 'What Do Children Want from a Social Robot? Toward Gratifications Measures for Child-Robot Interaction', in *2019 28th IEEE International Conference on Robot and Human Interactive Communication (RO-MAN)*, Oct. 2019, pp. 1–8. doi: 10.1109/RO-MAN46459.2019.8956319.

[44]   A.-M. Velentza, N. Fachantidis, and I. Lefkos, 'Learn with surprize from a robot professor', *Comput. Educ.*, vol. 173, p. 104272, Nov. 2021, doi: 10.1016/j.compedu.2021.104272.

[45]   C. N. Majova, 'Secondary school learners' attitudes towards sex education', Thesis, 2002. Accessed: Feb. 09, 2023. [Online]. Available: http://uzspace.unizulu.ac.za/xmlui/handle/10530/164

[46] A.-M. Velentza, N. Fachantidis, and I. Lefkos, 'Human or Robot University Tutor? Future Teachers' Attitudes and Learning Outcomes', in *2021 30th IEEE International Conference on Robot Human Interactive Communication (RO-MAN)*, Dec. 2021, pp. 236–242. doi: 10.1109/RO-MAN50785.2021.9515521.

[47] 'ELAN (Version 5.9) [Computer software]. (2020). Nijmegen: Max Planck Institute for Psycholinguistics. Retrieved from https://archive.mpi.nl/tla/elan"'.

[48] B. Streubel, C. Gunzenhauser, G. Grosse, and H. Saalbach, 'Emotion-specific vocabulary and its contribution to emotion understanding in 4- to 9-year-old children', *J. Exp. Child Psychol.*, vol. 193, p. 104790, May 2020, doi: 10.1016/j.jecp.2019.104790.